\documentclass[letterpaper, 10 pt, conference]{ieeeconf}  

\IEEEoverridecommandlockouts                              

\usepackage[usenames,dvipsnames]{xcolor}
\usepackage[colorinlistoftodos]{todonotes}
\usepackage{amsmath,amssymb}
\usepackage{algorithmic}
\usepackage{graphics}%
\graphicspath{ {Images/} }
\usepackage{url}

\title{\LARGE \bf
Quantifying the Reality Gap in Robotic Manipulation Tasks
}
\author{Jack Collins$^{1,2}$, David Howard$^{2}$ and J\"urgen Leitner$^{1,3}$%
\thanks{This research was supported by a Data61 PhD Scholarship.
This research was further supported by the Australian Research Council Centre of Excellence for Robotic Vision (project number CE140100016).}
\thanks{$^{1}$ Queensland University of Technology (QUT), Brisbane, Australia}%
\thanks{$^{2}$ Data61/CSIRO, Brisbane, Australia}%
\thanks{$^{3}$ Australian Centre for Robotic Vision (ACRV)}%
}

\begin{document}

\maketitle
\thispagestyle{empty}
\pagestyle{empty}

\begin{abstract}
We quantify the accuracy of various simulators compared to a real world robotic reaching and interaction task.
Simulators are used in robotics to design solutions for real world hardware without the need for physical access.
The `reality gap' prevents solutions developed or learnt in simulation from performing well, or at at all, when transferred to real-world hardware.
Making use of a Kinova robotic manipulator and a motion capture system, we record a ground truth enabling comparisons with various simulators, and present quantitative data for various manipulation-oriented robotic tasks.  We show the relative strengths and weaknesses of numerous contemporary simulators, highlighting areas of significant discrepancy, and assisting researchers in the field in their selection of appropriate simulators for their use cases. All code and parameter listings are publicly available from: https://bitbucket.csiro.au/scm/\textasciitilde{}col549/quantifying-the-reality-gap-in-robotic-manipulation-tasks.git.


\end{abstract}

\section{Introduction}

Simulators are widely used in the robotics community as they allow for real world systems to be quickly and cheaply prototyped without the need for physical access to hardware. Although used throughout robotics as a whole, simulators are particularly amenable to usage in robotic learning research.

The advent of data-hungry Deep Learning approaches, particularly Reinforcement Learning, heavily employ simulation to overcome the high costs intrinsic to repeated real-world data collection experiments, as well as obviating potential damage to expensive hardware during the early stages of learning. Simulated environments harness increasingly powerful and ubiquitous compute resources to cheaply and quickly generate synthetic data to accelerate the learning process. The use of simulation over reality carries numerous advantages, namely:

\begin{itemize}
\item No wear or damage to real-world hardware.
\item Many instantiations of a simulation can run in parallel;
\item (Often) Faster than real-time operation;
\item Instant access to robots without having to purchase; and
\item Human intervention is not required;
\end{itemize}

However, these benefits come with downsides; primarily that there are discrepancies between simulations and the real world brought about by the necessity to abstract, approximate, or remove certain physical phenomena, which prevents control systems created in simulation from performing to the same standard in reality. Learning-based approaches are known to exploit situations and achieve goals which are simulated artefacts and not realistically plausible in the real world \cite{lehman2018surprising}, further complicating the transfer to real world robotic applications. This disparity is a prominent issue with recent efforts in sim-to-real learning~\cite{Sunderhauf2018DL, ZhangModularPolicies}, as well as in Evolutionary Robotics (where simulations are crucial to speeding up these iterative, population-based algorithms), where it is referred to as `Reality Gap' \cite{jakobi1995noise}. Gaps mainly relate to actuators (i.e.~torque characteristics, gear backlash, ...), sensors (i.e.~sensor noise, latencies, and faults), temporal dynamics, and the physics that govern interactions between robots and objects in their environment (i.e.~deformable objects, fluid dynamics, ...).  

\begin{figure}[t]
    \vspace{2mm}
	\centering
	\includegraphics[width=\linewidth]{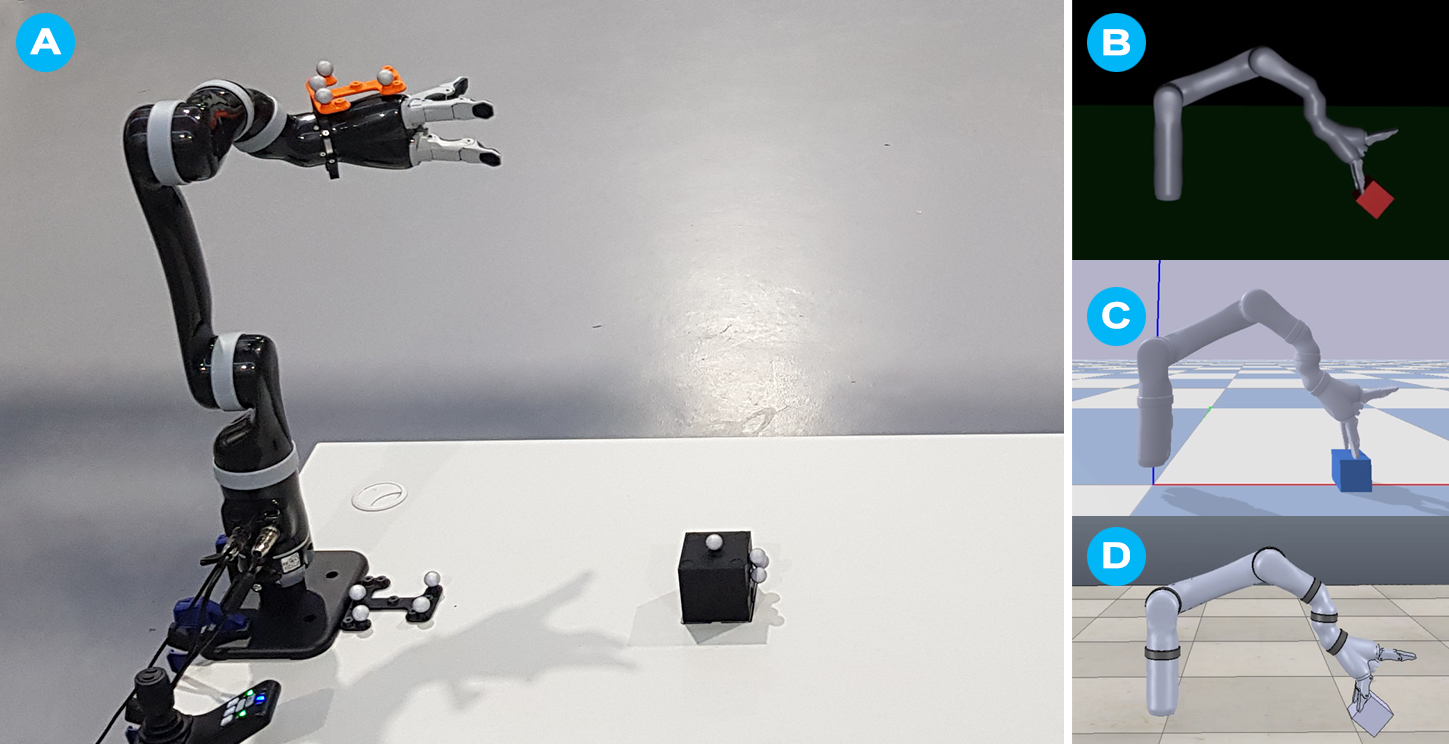}
	\caption{Real-World and simulated environments with the Kinova arm and cube visualised. (A) Real-world setup with tracking markers attached to dynamic elements; (B) the MuJoCo environment; (C) the PyBullet environment; (D) the V-Rep environment.}
	\label{VisualAbstract}
\end{figure}

Here we focus on reality gaps found in robotic grasping, which is selected as a `grand challenge' that is actively harnessing simulation-based learning \cite{levine2016end, lenz2015deep}, and is relevant to a vast swathe of application domains, ranging from industrial assembly to assisted living. From a simulation perspective, grasping is particularly challenging as interactions frequently occur between the robot and objects in its environment, which are rarely captured with any real veracity.

With a growing selection of physics engines and simulation environments available to researchers, the 'correct' combination of simulator/physics for a given task is becoming harder to ascertain. It is also becoming more and more important to know where these gaps exist, and how large they are, as a precursor to overcoming them such that simulation and reality more seamlessly meld. It is therefore timely and important to know how accurate these simulators are when performing various tasks, both to appropriately select a simulator for a particular research endeavour.

In this paper, we attempt to {\em quantify the reality gap}. We do this for a range of robotic manipulation experiments performed by a real 6DOF Kinova Mico2 arm. We simulate the same scenarios across a range of popular simulators and physics engines, and compare the data from the simulation runs to the real movements of the manipulator as recorded by a highly accurate motion capture setup, which we use as a ground truth (Fig.~\ref{VisualAbstract}).

The question we endeavour to answer is; \textit{to what accuracy can a range of popular robotic simulators replicate real world manipulation-related tasks?}  In particular, we ask;

\begin{itemize}
\item What are the differences between the chosen physics engines when simulating the same scenario?
\item Are there specific \textit{types} of interactions that some simulators can accurately model, compared to other simulators we test? 
\end{itemize}

We provide a detailed statistical analysis of these simulators when approximating movements of the real Kinova arm. Results quantify the disparity between the trajectory of the simulated Kinova arm and the real-world arm, and highlight that certain movements of the arm are more susceptible to misrepresentation in the simulator.  

Our work provides novel contributions to several fields of research in robotics, Deep Machine Learning, Evolutionary Robotics and Manipulation to name a few. We supply strong evidence to measure the accuracy of various simulators and physics engines when compared to a real-world ground truth. Additionally, we provide evidence that simulators are able to model the control and kinematics of manipulators accurately, but the dynamic interactions of a simulation remain unsolved. Our research is set to assist fellow researchers in the selection of simulators for their manipulation tasks.


\section{Related Work}\label{Related Work}

Robotics is an embodied discipline focused on building systems that act in the physical world. However, for numerous reasons highlighted in Section I, simulation is a key tool to many successful robotic engineering and integration efforts.  Simulation is fast, cheap, and allows for rapid prototyping and iteration over the composition and control of a robotic system.  These benefits are perhaps most strongly felt when learning is used, due to the data-hungry nature of many contemporary learning approaches.  Because simulators necessarily abstract various features (e.g., sensory delays, actuator slop), away from the physical reality, there exists a gap between what is simulated and how the final system performs in the real world.
Of course, we can in some situations learn directly on real hardware, however this requires sophisticated learning testbeds ~\cite{Howard2015AControllers,pinto2016supersizing,Heijnen2017AHardware} and, depending on the amount of data required, may be prohibitive in terms of required resources~\cite{levine2016learning}. Here we focus on simulated efforts to learn.

\subsection{Bridging the Reality Gap}
This `reality gap' is of increasing importance, as current deep learning approaches require a significant amount of data to achieve acceptable performance. Although increased computing power has narrowed this gap by facilitating more complex, high-fidelity simulations~\cite{Collins2018TowardsComponents}, the issue is as yet unsolved. 

Domain randomisation is a popular technique in robotic vision, whereby a trained model is subjected to randomised inputs (i.e., colour, shading, rendering, camera position, etc.)~\cite{James2017TransferringTask, Borrego2018AGazebo}.  Tobin~\textit{et al.}~\cite{Tobin2017DomainWorld} employ visual randomisation to teach a manipulator the 3D object position in simulation with a reasonable transfer to the real-world. Such approaches trace their lineage back to the first mention of the `reality gap', in the context of evolutionary robotics, which found success by introducing sensory noise into a simulator to discourage overspecialisation to simulated artefacts~\cite{jakobi1995noise}.

This sim-to-real transfer problem has recently been tackled by numerous research groups. Earlier approaches mainly highlight the issues around this transfer~\cite{Zhang2015TowardsControl}, with more recent efforts proposing solutions, including domain adaptation and Generative Adversarial Networks (GAN) which requires both real world and simulated data~\cite{ZhangModularPolicies}.  Results showing the early promise of these techniques --- using domain adaptation, Bousmalis~\textit{et al.}~\cite{Bousmalis2017UsingGrasping} were able to achieve a success rate for real world grasping trained in simulation of $76.7\%$ on a dataset of unseen objects.

An alternative method to randomisation is the optimisation of the simulated environments, with the goal to emulate the real world better. This approach requires real world data for the simulator to be able to fit to the real world observations, making it robot and application specific~\cite{Zagal2004BackRobotics, 10.1007/978-3-642-02921-9_11}.

There are several methods originating in Evolutionary Robotics that focus on grading a simulation based on the confidence of its prediction in an attempt to avoid poorly simulated scenarios. One such method implemented by Koos~\textit{et al.}~\cite{Koos2010CrossingControllers} offers a multi-objective approach that optimises both the fitness and the transferability of controllers. The transferability of a controller is evaluated using a surrogate model generated from data collected from controllers previously transferred to the test robot. Mouret~\textit{et al.}~\cite{Mouret201720Gap} state that a promising idea to cross the reality gap is to teach the limits of the simulator to a supervised learning algorithm with access to a real robot. This is then used to provide an accuracy prediction for simulated controllers. They report increased performance of the generated controllers. These scoring methods reduce the Reality Gap, but do so by limiting the simulator to predicting only things that it can accurately calculate, which reduces the applicability of the approach. They also require real-world data recorded directly from the platform to improve the simulation. 


Other approaches employ multiple simulators to overcome the biases from a single simulator. Boeing~\textit{et al.}~\cite{Boeing2012LeveragingGap} created the Physics Abstraction Layer (PAL), a unified interface between multiple physics engines and successfully evolved a PID controller for an Autonomous Underwater Vehicle. More recently Eaton~\textit{et al.}~\cite{Eaton2016BridgingRobot} evolved behaviour for a Nao robot using first the V-Rep simulator and then for successful controllers the Webots simulator to remove controllers that were exploiting unrealistic scenarios. The evolved controllers showed improved real-world performance after a small amount of human intervention to rectify an instability of the humanoid robot.  



\subsection{Physics Engines}

There are many physics engines targeting such diverse fields as gaming, movie effects, and robotics. 
Physics engines are created to model real-world physical properties in computer simulations with properties such as gravity, friction and contacts typically computed. These models are a simplification of the real-world, to compute a reasonable approximation within a restricted time and resource budget. 

Reviews of physics engines in the past have proven many times over that no one engine is capable of modelling all scenarios. Boeing~\textit{et al.}~\cite{Boeing2007EvaluationSystems} compared PhysX, Bullet, JigLib, Newton, Open Dynamics Engine (ODE), Tokamak and True Axis; they reported that Bullet performed best overall however no physics engine was best at all tasks. Chung~\textit{et al.}~\cite{Chung2016PredictableEngines} likewise found when testing Bullet, Dynamic Animation and Robotics Toolkit (DART), MuJoCo, and ODE, that no one engine performed better at all tasks, stating that for different tasks and different conditions a different physics engine was found to be better. These findings are further corroborated by Gonzalez-Badillo~\textit{et al.}~\cite{Gonzalez-Badillo2014AAssembly}, who showed that PhysX performs better than Bullet for non-complex geometries but is unable to simulate more complex geometries to the same degree as Bullet. 

One aim of our research is to provide a comprehensive study focused specifically around manipulation tasks, which we believe will be useful to the research community given the ongoing popularity of `learning to grasp'. Although several other researchers have evaluated physics engines, varying complexity and tasks~\cite{Fabry2016InteractiveRobotics, Gonzalez-Badillo2014AAssembly, Erez2015SimulationPhysX}, little research has been done on comparing real-world data to simulated data to draw conclusions as to the accuracy of physics engines and simulators. To our knowledge this is the first research that compares a highly accurate motion capture baseline with modern physics engines and simulators for real-world robot interaction tasks.



\subsection{Simulation Selection}

The list of robotic simulators is long, with many niche areas targeted by specific simulators. We are interested in robotic manipulation and looking for mature, well maintained simulators with active communities and good documentation practices to facilitate the development of robotics research. Additionally, we wanted to find a collection of simulators that provided a common programming language interface whilst also providing access to the Robot Operating System (ROS). 
We were left with the following: V-Rep, MuJoCo and PyBullet. These simulators expose the following 5 physics engines: Bullet, ODE, Vortex, Newton and Mujoco. This range of physics engines and simulators is attractive due to the range and crossover that the simulators afford whilst also providing a mature user interface.




\section{Method}\label{Method}	

The setup consists of a Kinova Mico2 6DOF arm with an attached KG-3 gripper.  The arm sits on a table, next to a manipulable cube. Simulators use the official Kinova URDF file.  In all cases, the Kinova arm is controlled using joint velocities as it allows higher fidelity over the position controller, and avoids the issue of a simulators position controller interfering with the sent motion commands. A basic proportional controller updated at $5$Hz is used to control the joints as this rate was feasible for all selected simulators.  We perform sets of identical movements, such that each movement happens in each simulator, and on the real arm, and record the results in each case.


\subsection{Real World Ground Truth}

\begin{figure}[t]
    \vspace{2mm}
	\centering
	\includegraphics[width=\linewidth]{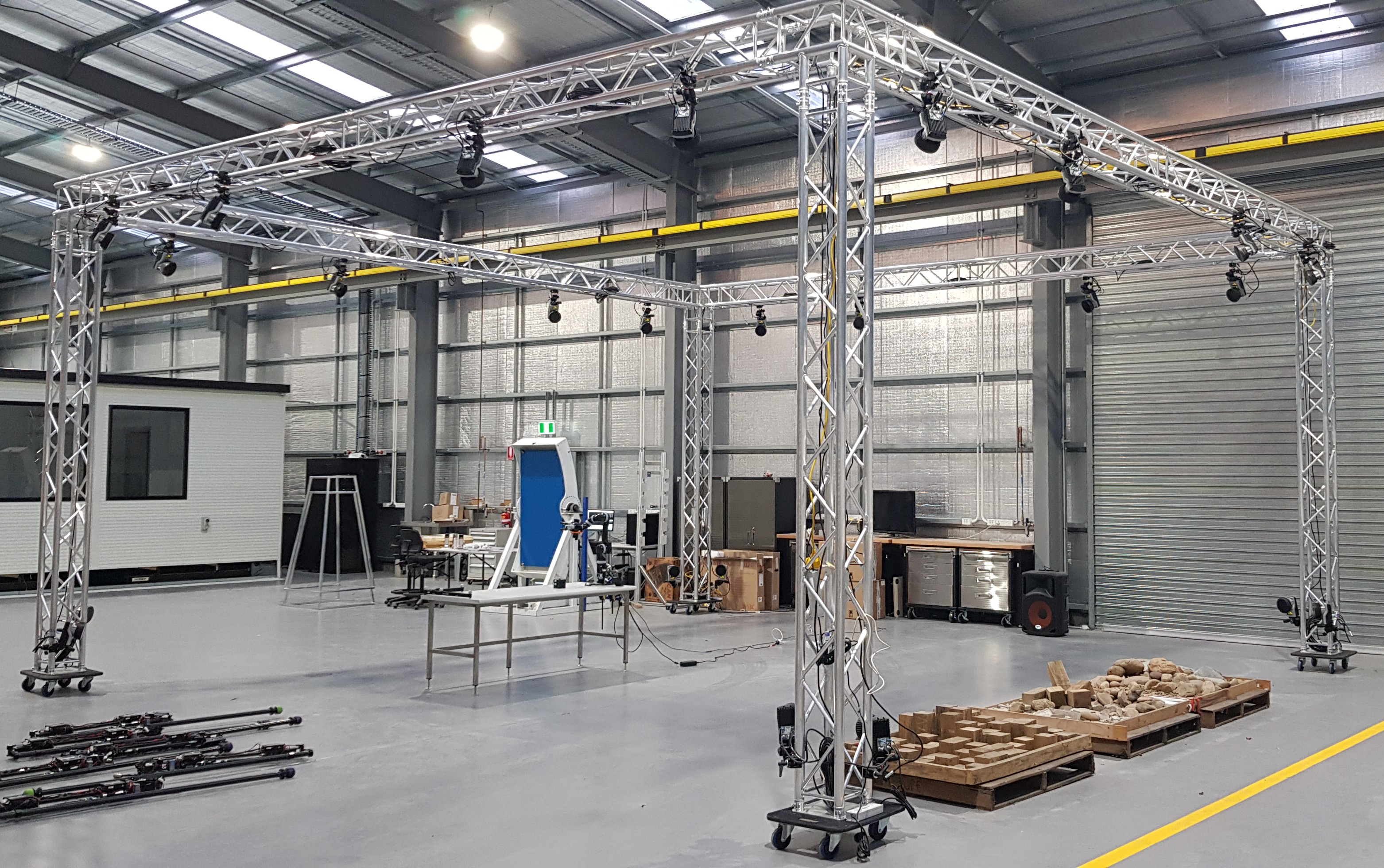}%
	
	\caption{Motion Capture System: 24 cameras fixed on a $8\times8\times4$ metre gantry records marker position at $100Hz$  to within $1$ millimetre accuracy.}
    \label{MotionCapture}
\end{figure}

\begin{figure}[h]
  \centering
  \includegraphics[width=\linewidth]{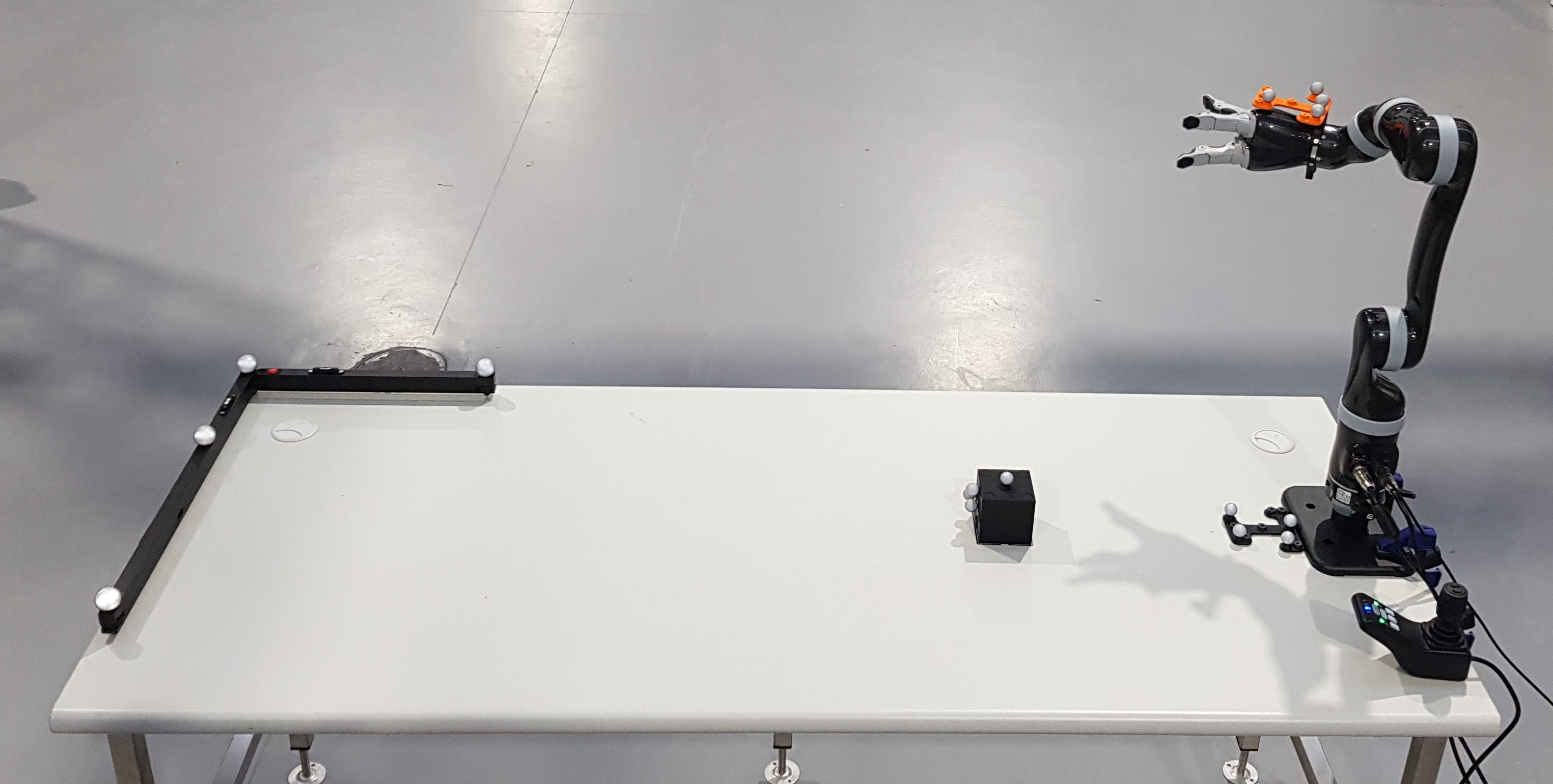}
  \caption{The real-world setup of the 6DOF Kinova Mico2 Arm with tracking markers attached to wrist via a rigid marker base. 
  On the opposite end of the table sits the L-frame which acts as the global co-ordinate system. The cube with 4 markers attached is also visible. }
  \label{Configuration}
\end{figure}

A Qualisys motion capture system utilising 24 cameras mounted to a $8\times8\times4$ metre gantry records the real-world data (Fig.~\ref{MotionCapture}). A $0.75\times1.8$ metre table with a laminated table top acts as the ground plane for all experiments.
Four tracking markers were attached to the wrist of the Kinova arm using a 3D printed mount and rigid marker base.  Another rigid base with 4 markers lies on the table top flush against the bracket supporting the arm. The global coordinate frame sits on the opposite end of the table, out of reach of the arm.
A 3D printed cube, made from ABS plastic, sits on top of the table.  The cube is $7.5$cm per side, with a weight of $88.4$ grams, including 4 tracking markers (Fig.~\ref{Configuration}).

We recorded 6DOF poses -- x, y and z positions and orientation as Euler angles -- of the rigid bases with offset rotations and translations for the co-ordinate systems. The co-ordinate system of the wrist mounted base was set to be at the centre of the wrist, analogue to the simulations. The co-ordinate system at the centre of the cube also followed the simulators XYZ frame. 
The base marker was used as the global co-ordinate system for both the cube and the wrist tracking, allowing for comparative results between simulation and motion tracking.

Control of the Kinova arm was through ROS using the official Kinova package, which supplies joint rotations in degrees and allows for joint velocity commands to be sent at a rate of $100Hz$. Using the same proportional controller and actions as generated in simulation, scenarios were able to be run in python using the ROS interface\footnote{All code and parameter listings are publicly available from: https://bitbucket.csiro.au/scm/\textasciitilde{}col549/quantifying-the-reality-gap-in-robotic-manipulation-tasks.git}.


\subsection{Simulation}


Three leading simulators are compared in our experiments: V-Rep \cite{Rohmer2013V-REP:Framework}, Mujoco \cite{Todorov2012MuJoCo:Control} and PyBullet \cite{E.CoumansandY.Bai2016PybulletLearning}.
The Kinova arm was imported into each simulator's scene with the cube modelled as a primitive cuboid object.
The following points highlight the additional simulator specific changes required after importing the manipulator, with the only shared changes being the starting pose and the starting position (elevated $0.055$ metres to account for the Kinova base plate). Other general setup included modelling the weight ($0.0884$ kilograms), size ($0.075m^3$) and position (0.5,0,0.375) of the cube.
All other parameters of the simulations were kept to each simulator's defaults, unless otherwise stated. These include friction models, inertia properties, actuator settings, simulation step sizes, integrators, solvers, etc. The majority of settings are left to their default value as we want to see how well a generic scene can perform without the knowledge of an expert.

\subsubsection{V-Rep}
The scene was imported using V-Reps plug-in and saved as a .ttt binary file after creation. The joint settings were changed to ``Lock motor when target velocity is zero".
\subsubsection{PyBullet}
PyBullet's time step was explicitly fixed to the value of $0.01$ seconds. 
\subsubsection{MuJoCo}
The {\textit mujoco-py} Python wrapper maintained by OpenAI was used as the interface for MuJoCo. The URDF needed to be converted to an Extensible Markup Language (XML) file with MuJoCo modelling layout; this was done using the MuJoCo compile script. The actuator type and sensors needed to be added manually with the only altered parameter in the XML file being the $kv$ velocity feedback gain. The simulator time step of the simulation was set at $0.0001$ seconds as this provided a stable simulation. 

\section{Experiments and Results}\label{Experiments}
The experimentation is designed to assess the ability of robotic physics simulations to reproduce real-world scenarios. All experiments are repeated 20 times to ensure reproducible, unbiased results. Data collected for each experiment is limited to the 6DOF pose of the Kinova wrist joint and for one experiment the 6DOF pose of the cube.
There are three scenarios in total:
(a) beginning with a very basic robotic movement of one joint,
(b) moving onto more complex multi-joint movement tasks, and
(c) finally an interaction task where the robot arm is pushing the cube along the table.

The motion capture system once calibrated provides accuracy to within $1$ millimetre and records $100\%$ ``measured'' data without the need for interpolation.  We therefore consider the motion capture an accurate approximation of the ground truth in our real-world experiments, and the baseline we compare the simulators to.

\subsection{Scene 1: Single Joint Movement}
This experiment was designed to compare the control of a single joint of the Kinova arm.
For that (Joint \#2) rotates from a starting to a final pose, for a rotation of $100$ degrees, in about 6 seconds (i.e.~$120$ control cycles at $5Hz$). All other joints are controlled to the set rotation of $0$ degrees. 

\begin{figure}[tb]
	\centering
	\includegraphics[width=\linewidth]{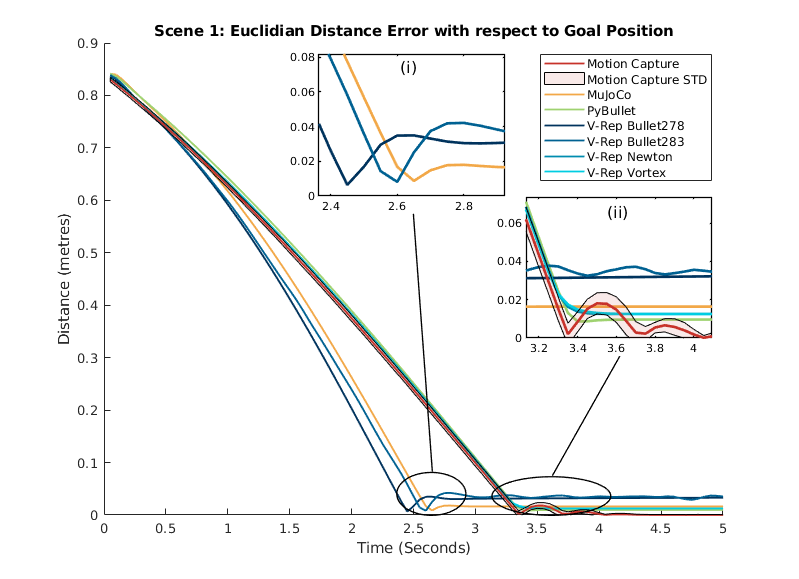}
	\caption{A single joint motion performed on the Kinova arm both in the real world and in simulators.
    Plotted is the mean Euclidean distance from the goal position, calculated from 20 runs. (i) and (ii) are areas of note within the plot.}
	\label{singleEuclidianError}
\end{figure}

Fig.~\ref{singleEuclidianError} presents the results for Scene 1 data as a graph where each plot is the 
euclidean distance error (Eq.~\ref{euclideanError}) plotted over time. Where $p_{x,y,z}$ are the mean position of the motion capture/physics engine at each time step and $g_{x,y,z}$ is the goal position of the wrist at $100$ degrees.\
\begin{equation}
e = \sqrt[]{(p_{x}-g_{x})^{2}+(p_{y}-g_{y})^{2}+(p_{z}-g_{z})^{2}}
\label{euclideanError}
\end{equation}

Most noticeable in Fig.~\ref{singleEuclidianError} is the lack of results for the ODE physics engine, this is due to the instability of the V-Rep simulation which appears to be caused by self-contacts of the Kinova arm model. The accumulated error for ODE as seen in Table~\ref{sumTable} proves the instability of the physics engine through the comparatively large value; this could not be rectified by tuning the parameters of the simulation. All plotted results begin at the same start position, with Vortex, Newton and PyBullet following the Motion Capture error most closely. This is quantified in Table~\ref{sumTable} where the accumulated error for Vortex, Newton and PyBullet is markedly lower then the other physics engines. 

The convergence of the physics engines to the goal position is also of note, as Bullet283 and Bullet278 arrive approximately $1$ second earlier than all other plots. Bullet283 and Bullet278 also oscillate noticeably before reaching their final state (Fig.~\ref{singleEuclidianError} (i)) which replicates the motion capture convergence as it too oscillates (Fig.~\ref{singleEuclidianError} (ii)).  The remaining physics engines show very little to no oscillation. Also included in the plot is the standard deviation of the motion capture system for comparison. Standard deviations for other plots are not displayed as the discrepancies in simulation are negligible. Finally, it appears that the motion capture is the only one to reach exactly $100$ degrees as no other plots reach the same final position.

\subsection{Scene 2: Multi Joint Movement}
\begin{figure*}[tbh]
  \centering
	\includegraphics[width=\linewidth]{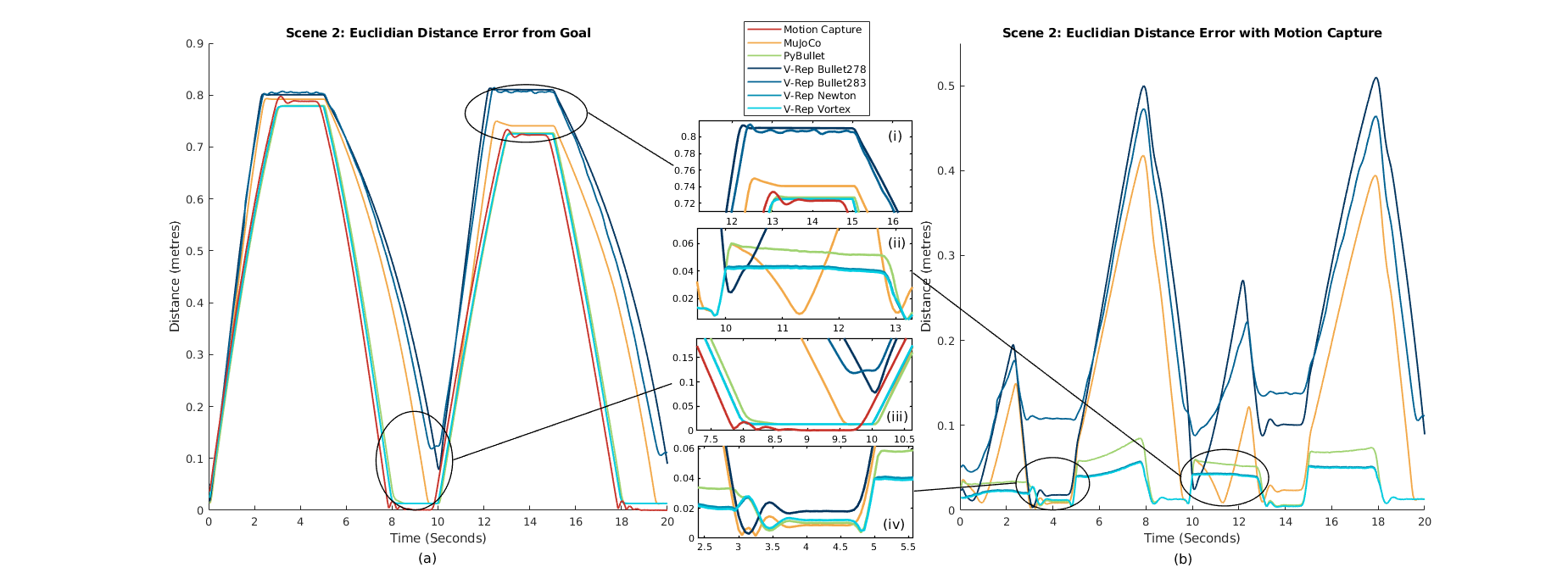}
	\caption{Plot of Scene 2 with the goal position set to the ground truth. Plotted lines are the euclidean distance from the goal position calculated using the mean of 20 results. (i) and (ii) are areas of note within the plot.}
	\label{doubleEuclidianError}
\end{figure*}

Scene 2 is a more complex scene where joints two and five are moved multiple times within $20$ seconds. Joint 2 is programmed to move between $0~|~90~|~0~|~90~|~0$ degrees and joint 5 moves between $0~|~90~|~0~|~-90~|~0$ degrees. Fig.~\ref{doubleEuclidianError} depicts the euclidean error plot (Eq.~\ref{euclideanError}) where $g_{x,y,z}$ for (a) is the final goal position of the wrist and (b) is the equivalent time step motion capture position. Newton and Vortex follow the motion capture path closely with an accumulated error of $\pm5.5-6$ metres, while PyBullet also has a low accumulated error of $\pm7$ metres. MuJoCo, Bullet283 and Bullet278  model the motion capture closer between $0-5$ seconds and $10-15$seconds, this is due to the arms moving with gravity towards the goal state and then during the error-full periods slowly moving aginst gravity and accruing more error. When moving against gravity only MuJoCo is able to reach the final position before changing trajectory. Some of the simulators (i.e. Mujoco and Bullet283) also generate the oscillation seen by the motion capture as the proportional controller attempts to correct the rotation, although none are able to imitate the exact motion of real robot. 

\subsection{Scene 3: Interaction with the cube}
\begin{figure*}[tbh]
	\centering
 	\includegraphics[width=\linewidth]{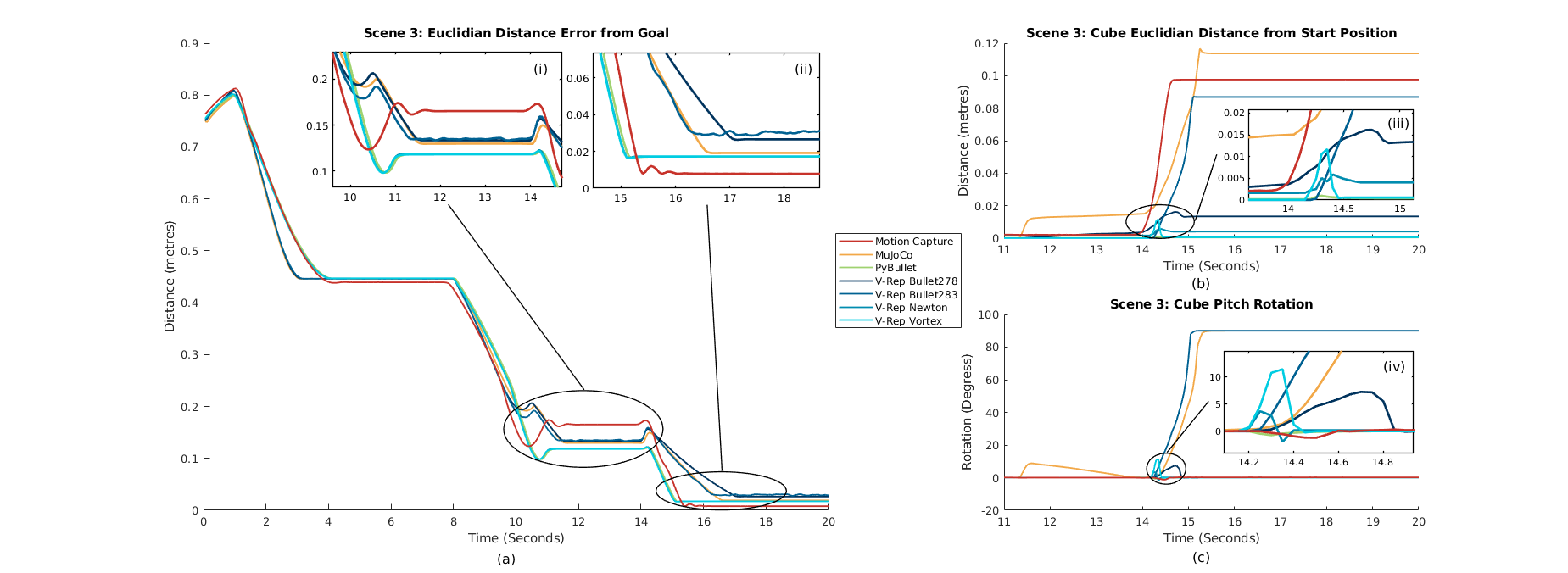}
	\caption{Three plots of Scene 3: (a) plotted lines represent the wrist's euclidean distance from the final goal position calculated using the mean of 20 results; (b) the x-axis represents the start position of the cube with lines the cubes euclidean distance away; and (c) the pitch of the cube (where pitch is the rotation around the y-axis).(i-iv) are areas of note within the plot.}
	\label{cubeError}
\end{figure*}

The most complex scene where joints two, three and four move in a sequence and push a cube along a flat plane within 20 seconds. There are three phases of this scene, the first two position the arm to make contact with the cube and the third initiates the contact and pushes the cube. This scene tests both the control and the physics of the system through the movement of the Kinova arm and the interaction with the cube. Fig.~\ref{cubeError} shows three plots: (i) the euclidean distance error (Eq.~\ref{euclideanError}) where $g_{x,y,z}$ is the goal position of the wrist; (ii) the euclidean distance error of the cube (Eq.~\ref{euclideanError}) where $g_{x,y,z}$ is the start position of the cube (i.e. x:0.5, y:0, z:0.0375); and (iii) the rotation of the cube around the y-axis. 
The first plot shows that no physics engine outperforms any other by a distinguishable margin. This is reinforced by the results in Table~\ref{sumTable},  where the accumulated error for Newton, Vortex and Pybullet are approximately $\pm45$ metres. It also appears that the motion capture is the closest to reach the goal state, however all plots settle close to the x-axis.
The second plot shows the physical interaction between two rigid objects. The greatest displacement is made by MuJoCo followed by the Motion Capture, Bullet283, and then Bullet278. Vortex has very little displacement as the cube makes minimal contact with the Kinova gripper due to the large error seen in the first plot at $15$ seconds. Pybullet does not interact with the cube at all, with the gripper moving over the cube. Mujoco is the first to interact with the cube and does so early at about $11.4$ seconds whereas all other physics engines begin at about $14$ seconds; this is at the conclusion of the previous phase designed to get the gripper in a position to interact with the cube.
The final plot shows the pitch of the cube and this is important due to the discrepancies between the physics engines and the ground truth. The plot clearly shows that both Bullet283 and MuJoCo knock the cube in such a way that it rotates 90 degrees. The same movement in reality moves the cube forward with only the smallest amount of discernible rotation. The only physics engine which is able to match the lack of rotation is PyBullet and that is due to it not interacting with the gripper at all. This is particularly relevant given our focus on robotic manipulation, which would benefit greatly from reasonable modelling of these multi-body interactions.

The crossover between physics engines shows PyBullets implementation of Bullet and V-Rep's two implementations of Bullet. The total column from Table~\ref{sumTable} shows a vast difference between the two simulators, with PyBullet drastically better at simulating the chosen scenes with a cumulative position error of $\pm45.5$ metres while V-Rep's implementations of Bullet both attained $\pm131$ metres. This could be due to several effects, the first being the default values being set differently (i.e. we set the timestep of PyBullet to be $0.01$ seconds while V-Rep generically uses $0.05$ seconds), the second could be the underlying implementation between the V-Rep simulator and the physics engine at a scale inaccessible to the user.

For the control of the manipulator Newton, Bullet (PyBullet implementation) and Vortex were considerably and consistently better. For interaction between objects there was no physics engine that modelled the collision well. In reality the cube moved a total of $0.0975$ metres in x,y and z with a change in rotation of roll: $0.01$, pitch: $0.11$ and yaw: $0.89$ degrees. The physics engines that were closest to modelling position (Mujoco: $0.1137$ metres and Bullet283: $0.0869$ metres) had incorrect rotations (MuJoCo and Bullet283 pitch: $90$ degrees) and those that had similar real-world rotations had minimal positional movement ($<0.0134$ metres). 

By stitching together physics engines for discrete periods within a simulation it is believed that a model capable of further reducing the reality gap can be generated. This is backed up by the results which show for the control of the manipulator we should select Newton, PyBullet and Vortex to model the kinematics, without using the results of the remaining physics engines. The control segments could then be combined to the results of MuJoCo, Bullet283 and Bullet278 for the period of interaction with the cube. By populating periods in the simulation timeline with only the optimal performing physics engine the resulting simulation should display a closer realisation of reality. 


\begin{table}[t!]
\caption {Accumulated (over timesteps) Euclidean Error (m) compared to the Ground Truth}
\begin{tabular}{|l|c|c|c|c|}
\hline
\textbf{}                  & {\textbf{1 Joint}} & {\textbf{2 Joints}} & \textbf{Cube} & {\textbf{Total}} \\ \hline
\textbf{MuJoCo}            & 24.237                               & 49.430                               & 23.471                               & 97.138                             \\ \hline
\textbf{PyBullet}          & \textbf{18.429}                               & 7.000                                & \textbf{20.084}                               & 45.513                             \\ \hline
\textbf{V-Rep} (Bullet2.78) & 27.412                               & 81.166                               & 25.034                               & 133.611                            \\ \hline
\textbf{V-Rep} (Bullet2.83)& 26.698                               & 80.004                               & 25.215                               & 131.916                            \\ \hline
\textbf{V-Rep} (Newton)     & 18.810                               & \textbf{5.579}                                & 21.069                               & 45.458                             \\ \hline
\textbf{V-Rep} (Vortex)    & 18.887                               & 5.664                                & 21.130                               & 45.680                             \\ \hline
\textbf{V-Rep} (ODE)        & 1.31e+17                              & 1.19e+18                              & 1.88e+18                              & 3.20e+18                          \\ \hline
\end{tabular}
\label{sumTable} 
\end{table}

\section{Conclusion}\label{Conclusion}
We have demonstrated the ability of a range of physics engines to simulate a set of manipulation tasks. Using a motion capture system as the ground truth we record 6DOF pose of a robotic manipulator and directly compare it to simulated data collected from the MuJoCo, PyBullet and V-Rep simulators. The range of tasks test the kinematic and dynamic modelling capabilities of Bullet, Mujoco, Newton, ODE and Vortex. Contributions are both the quantified evidence of the capability of physics engines to model manipulation tasks and the analysis of the simulated and real-world data, including a highly accurate ground truth. 

We show the simulation of the kinematic model and control of manipulators is largely solved when compared to the real world, however there are considerable developments necessary for interactions between simulated objects. The physics behind contacts remains a complex problem that is difficult to replicate in simulated environments, and we suggest that a focus on such interactions will bring increasing benefits for the 'learning to grasp' community.

The results highlight the strengths and weaknesses of contemporary simulators with focus on discrepancies between the real-world ground truth. Our contributions will assist researches in the field in their selection of simulators.



\bibliographystyle{IEEEtran}
\bibliography{Mendeley,references}
\newpage

\end{document}